\documentclass[10pt, a4paper]{article}
\pdfoutput=1

\usepackage{lrec2022}

\usepackage[utf8]{inputenc}
\usepackage[T1]{fontenc}
\usepackage{CJK}

\usepackage{amsmath,amsfonts,amssymb,bm,mathtools}
\usepackage{booktabs}
\usepackage{hyperref}
\usepackage{url}
\usepackage{xcolor}
\usepackage{appendix}
\usepackage{multirow}
\usepackage{makecell}
\usepackage{paralist}
\usepackage{longtable}
\usepackage{etoolbox}

\def\arxiv{1}

\if\arxiv1
\usepackage[round]{natbib}
\else
\let\citet\cite
\let\citep\cite
\fi

\usepackage{titlesec}
\titleformat{\section}{\normalfont\large\bfseries\center}{\thesection.}{1em}{}
\titleformat{\subsection}{\normalfont\SmallTitleFont\bfseries\raggedright}{\thesubsection.}{1em}{}
\titleformat{\subsubsection}{\normalfont\normalsize\bfseries\raggedright}{\thesubsubsection.}{1em}{}
\renewcommand\thesection{\arabic{section}}
\renewcommand\thesubsection{\thesection.\arabic{subsection}}
\renewcommand\thesubsubsection{\thesubsection.\arabic{subsubsection}}

\title{CVSS Corpus and Massively Multilingual Speech-to-Speech Translation}

\name{Ye Jia,~~Michelle Tadmor Ramanovich,~~Quan Wang,~~Heiga Zen} 

\address{Google Research\\
         jiaye@google.com\\}

\abstract{
We introduce \emph{CVSS}, a massively multilingual-to-English speech-to-speech translation (S2ST) corpus, covering sentence-level parallel S2ST pairs from 21 languages into English. CVSS is derived from the
Common Voice \citep{ardila2020common} speech corpus and the CoVoST 2 \citep{wang2020covost2} speech-to-text translation (ST) corpus,
by synthesizing the translation text from CoVoST~2 into speech using  state-of-the-art TTS systems.
Two versions of translation speech in English are provided: 1) \emph{CVSS\nobreakdash-C}: All the translation speech is in a single high-quality canonical voice; 2) \emph{CVSS-T}: The translation speech is in voices transferred from the corresponding source speech.
In addition, CVSS provides normalized translation text which matches the pronunciation in the translation speech.
On each version of CVSS, we built baseline multilingual direct S2ST models and cascade S2ST models, verifying the effectiveness of the corpus. To build strong cascade S2ST baselines, we trained an ST model on CoVoST~2, which outperforms the previous state-of-the-art trained on the corpus without extra data by 5.8 BLEU. Nevertheless, the performance of the direct S2ST models approaches the strong cascade baselines when trained from scratch, and with only 0.1 or 0.7 BLEU difference on ASR transcribed translation when initialized from matching ST models.

\vspace{1.2ex}

\Keywords{speech-to-speech translation, speech-to-text translation, multilingual, cross-lingual voice transferring}
}

\begin{document}

\maketitleabstract

\section{Introduction}

Speech-to-speech translation (S2ST) is an important means for breaking down the communication barriers between people speaking different languages. Conventionally, S2ST systems are built with a cascade of automatic speech recognition (ASR), text-to-text machine translation (MT), and text-to-speech (TTS) synthesis sub-systems, which are text-centric. Recently, work on S2ST without relying on intermediate text representation 
are emerging, such as end-to-end direct S2ST \citep{jia2019direct,kano2021transformer,jia2021translatotron} and cascade S2ST based on discrete speech representation \citep{tjandra2019speech,zhang2020uwspeech,lee2021direct,ma2021direct,lee2021textless}. 
However, as of today, publicly available corpora directly suitable for such research are extremely limited (see Table~\ref{tbl:corpora}).

In this paper, we introduce \emph{CVSS}, a \textbf{C}ommon \textbf{V}oice-based \textbf{S}peech-to-\textbf{S}peech translation corpus. 
CVSS is directly derived from the CoVoST 2 ST corpus, which is further derived from the Common Voice speech corpus.
CVSS provides sentence-level parallel speech-to-speech translation pairs from 21 languages into English, namely from Arabic (ar), Catalan (ca), Welsh (cy), German (de), Estonian (et), Spanish (es), Persian (fa), French (fr), Indonesian (id), Italian (it), Japanese (ja), Latvian (lv), Mongolian (mn), Dutch (nl), Portuguese (pt), Russian (ru), Slovenian (sl), Swedish (sv), Tamil (ta), Turkish (tr), and Chinese (zh).
The source speech in these 21 languages is crowd-sourced human volunteer recordings from the Common Voice project, totalling 1153 hours.
Two versions of translation speech in English are provided for all the source speech, both are synthesized using state-of-the-art TTS systems,
with each version providing unique values not existing in other public S2ST corpora:
\begin{itemize}
    \item \emph{CVSS-C}: All the translation speech is in a single canonical speaker's voice, totalling 719 hours.
    Despite being synthetic, the speech is highly natural, clean, and consistent in speaking style.
    These properties ease the modeling of the target speech and enable trained models to produce high quality translation speech suitable for general user-facing applications.
    \item \emph{CVSS-T}:  The translation speech is in voices transferred from the corresponding source speech, totalling 784 hours. Each S2ST pair has a similar voice on the two sides despite being in different languages, making this dataset suitable for building models where voice preservation during speech translation \citep{jia2021translatotron} is desired.
\end{itemize}
Together with the source speech, the two S2ST datasets contain 1,872 and 1,937 hours of speech, respectively.
In addition to translation speech, CVSS also provides normalized translation text matching the pronunciation in the translation speech (e.g. on numbers, currencies, acronyms, etc.), which can benefit both model training as well as evaluation.

Unlike existing corpora of simultaneous interpretation, e.g. VoxPopuli \citep{wang2021voxpopuli} and STC \citep{shimizu2014collection}, the target speech in CVSS is translation instead of interpretation. As a comparison, translation is typically verbatim and exact, while interpretation is typically summarizing and often drops less important details; there is also more linguistic variation and disfluencies in interpretation
\citep{he2016interpretese,shimizu2013constructing,wang2021voxpopuli}.

CVSS is released under the very permissive Creative Commons Attribution 4.0 International (CC BY 4.0) license. %
It can be freely downloaded online.\footnote{\url{https://github.com/google-research-datasets/cvss}}
\begin{table*}[t]
\centering
\footnotesize
\setlength{\tabcolsep}{0.3em}
\caption{Basic comparison of public S2ST corpora (including ST corpora used for S2ST).}
\vspace{1ex}
\begin{tabular}{llll@{\hskip -1.5em}r}
    \toprule
    Corpus & Languages & Source speech & Target speech & Total hours \\
    \midrule
    Fisher Es-En \citep{post2013improved} & es $\to$ en & Telephone conversations (8 kHz) & N/A & 127 \\
    STC\footnotemark{} \citep{shimizu2014collection}
    & en $\to$ ja & TED talks & Simultaneous interpretation & 31 \\
    MaSS \citep{boito2019mass}          &  8  (56 directions) & Read Bible & Read Bible & 150 \\
    VoxPopuli \citep{wang2021voxpopuli} & 15 (210 directions) & European Parliament speech & Simultaneous interpretation & 17.3K \\
    CVSS-C (this work) & X$\to$En (21 directions) & Read text & Synthetic (single voice) & 1.9K \\
    CVSS-T (this work) & X$\to$En (21 directions) & Read text & Synthetic (cloned voice) & 1.9K \\
    \bottomrule
\end{tabular}
\vspace{-2ex}
\label{tbl:corpora}
\end{table*}

On each version of CVSS, we built two baseline direct S2ST models (Translatotron \citep{jia2019direct} and Translatotron~2 \citep{jia2021translatotron}) and a baseline cascade S2ST model (ST $\to$ TTS).
To build strong cascade S2ST baselines, we trained an ST model on CoVoST~2, which 
outperforms the previous state-of-the-art trained on the corpus without using extra data by $+$5.8 average BLEU on all 21 language pairs, or $+$6.9 average BLEU on the 4 high-resource language pairs. Nevertheless, the performance of the Translatotron~2 direct S2ST model approaches the strong cascade baseline when trained from scratch,
and with only 0.1 or 0.7 BLEU difference on ASR transcribed translation when initialized from matching ST models. %
These results verified the effectiveness of both the CVSS corpus as well as the approach of direct S2ST. We hope the release of the CVSS corpus and the baselines we provide can help accelerate the research on direct S2ST.

\section{Related works}

Research on S2ST has progressed for over three decades since early efforts such as \citet{waibel1991janus}. However, publicly available corpora with parallel S2ST pairs are still extremely limited as of today. This is largely because until very recently, S2ST research has focused on the cascade approach, thus requiring separate ASR, MT, and TTS corpora. However, such corpora are not directly usable for building S2ST without relying on text representation.

Fisher Spanish-English ST corpus \citep{post2013improved} is the most widely used public corpus in recent S2ST works \citep{jia2019direct,zhang2020uwspeech,lee2021direct,ma2021direct}. It contains 127 hours of Spanish telephone conversations and corresponding English translation text. However, this corpus does not include translation speech, and all these works used their own versions of synthetic translation speech. The low sample rate (8~kHz) of the source speech also makes it less ideal for modern S2ST research.

VoxPopuli \citep{wang2021voxpopuli} is a recent large speech corpus originated from European Parliament event recordings. It includes 17.3k hours simultaneous oral interpretation in 15 languages, which is by far the largest S2ST corpus publicly available.
Because of the nature of oral interpretation, important information in the source speech can be missing in the interpretation. The variation in speakers' voices, recording conditions, and disfluencies in the  interpretation  pose additional challenges for S2ST modeling on this corpus.

MaSS \citep{boito2019mass} is a small corpus of Bible reading in 8 languages, with about 20 hours of speech per language.
STC \citep{shimizu2014collection} includes a small publicly available simultaneous interpretation corpus that interprets English TED Talks recordings into Japanese.\footnote[2]{The STC corpus includes en $\leftrightarrow$ ja simultaneous interpretation from multiple sources, but only a portion of the en~$\to$~ja direction has both the source and target speech available.}

A few recent works \citep{tjandra2019speech,kano2021transformer} used the BTEC corpus \citep{kikui2003creating,kikui2006comparative}, which is derived from a small hand-crafted MT corpus of phrases in the travel domain. This corpus is currently not available to be downloaded.
Similarly, a few other corpora with S2ST pairs are no longer publicly available, such as: EPIC \citep{bendazzoli2005approach}, containing 18 hours simultaneous interpretation among Italian, English and Spanish, originated from the European Parliament speech; CIAIR \citep{tohyama2004ciair}, containing 182 hours simultaneous interpretation between Japanese and English.

\paragraph{Synthetic vs Human-recorded}

Most of the above mentioned recent S2ST works use synthetic translation speech as training targets \citep{jia2019direct,tjandra2019speech,zhang2020uwspeech,kano2021transformer,lee2021direct,jia2021translatotron,ma2021direct}.
There are two primary motivations for doing so: 1) Collecting a large amount of synthetic speech is of much lower cost than collecting human recordings, in absence of a directly usable S2ST corpus; 2) Synthetic speech can be easier to model because of consistent voice, speaking style, and high cleanness.
\citet{jia2019direct,jia2021translatotron} showed that despite being training on synthetic speech, the trained S2ST models can produce translation speech in high naturalness.

A few works built S2ST models with real-world human recordings as training targets. Because large-scale human recordings usually have to be collected with multiple speakers in different recording conditions, these works have to introduce additional components for tackling the variation in speakers' voices, speaking styles, and recording conditions, etc. Such components are often trained with additional corpora. \citet{jia2019direct} used a speaker encoder separately trained with a speaker verification corpus, in order to model such variation. \citet{lee2021textless} used a speech normalizer separately trained with a synthetic speech normalization corpus and a speech quantizer separately trained with an unsupervised speech corpus, in order to eliminate such variation.

Besides collecting human speech or using synthetic speech to construct S2ST datasets, it is possible to mine S2ST data from existing multilingual untranscribed speech corpora. \citet{duquenne2021multimodal} showed a proof-of-concept of such an approach.

\paragraph{Text normalization}
The translation quality of S2ST is typically evaluated by measuring BLEU \citep{papineni2002bleu} between reference translation text and ASR transcription of the translation speech \citep{jia2019direct}. %
Because ASR usually outputs with minimal punctuation and case support\footnote{Particularly, most of recent S2ST works used ASR models trained on the LibriSpeech corpus \citep{panayotov2015librispeech} for such evaluation. LibriSpeech corpus provides text in uppercase without punctuation marks.},
such evaluation typically computes BLEU case-insensitively and ignores punctuation marks. In addition, a few works, e.g. \citet{lee2021textless}, further normalize special tokens such as numbers in reference text before computing BLEU. Such text normalization is not standardized, which makes the result comparison among different works difficult.
In CVSS, we provide normalized translation text matching the pronunciation in the target speech, which can be used for model training as well as help standardize the evaluation on this corpus.

\section{Source corpora}

CVSS is directly derived from the CoVoST 2 ST corpus, which is further derived from the Common Voice speech corpus.

\paragraph{Common Voice}\citep{ardila2020common} is a massively multilingual transcribed speech corpus designed for ASR. The speech in the corpus is crowdsourcing collected by volunteer contributors reading text content from Wikipedia and other text corpora. The size and the language coverage of the corpus keeps growing. The current release (version 7) consists of 11,192 hours of validated speech in 76 languages.

\paragraph{CoVoST 2} \citep{wang2020covost2} is a large-scale multilingual
ST corpus derived from Common Voice. It covers translation from 21 languages into English and from English into 15 languages. The source speech is directly from Common Voice version 4. The translation was collected from professional translators on the scripts from the Common Voice. The 21 X-En language pairs consist of 1,154 hours of speech in total.

\section{TTS models}

CVSS is constructed by synthesizing the translation text from CoVoST~2 into speech using two state-of-the-art TTS models. This section describes the two TTS models being used, both of which were trained on the LibriTTS corpus \citep{zen2019libritts}. 

\subsection{PnG NAT}
\label{sec:png-nat}

\begin{figure}[t]
  \centering
  \includegraphics[width=1.0\linewidth]{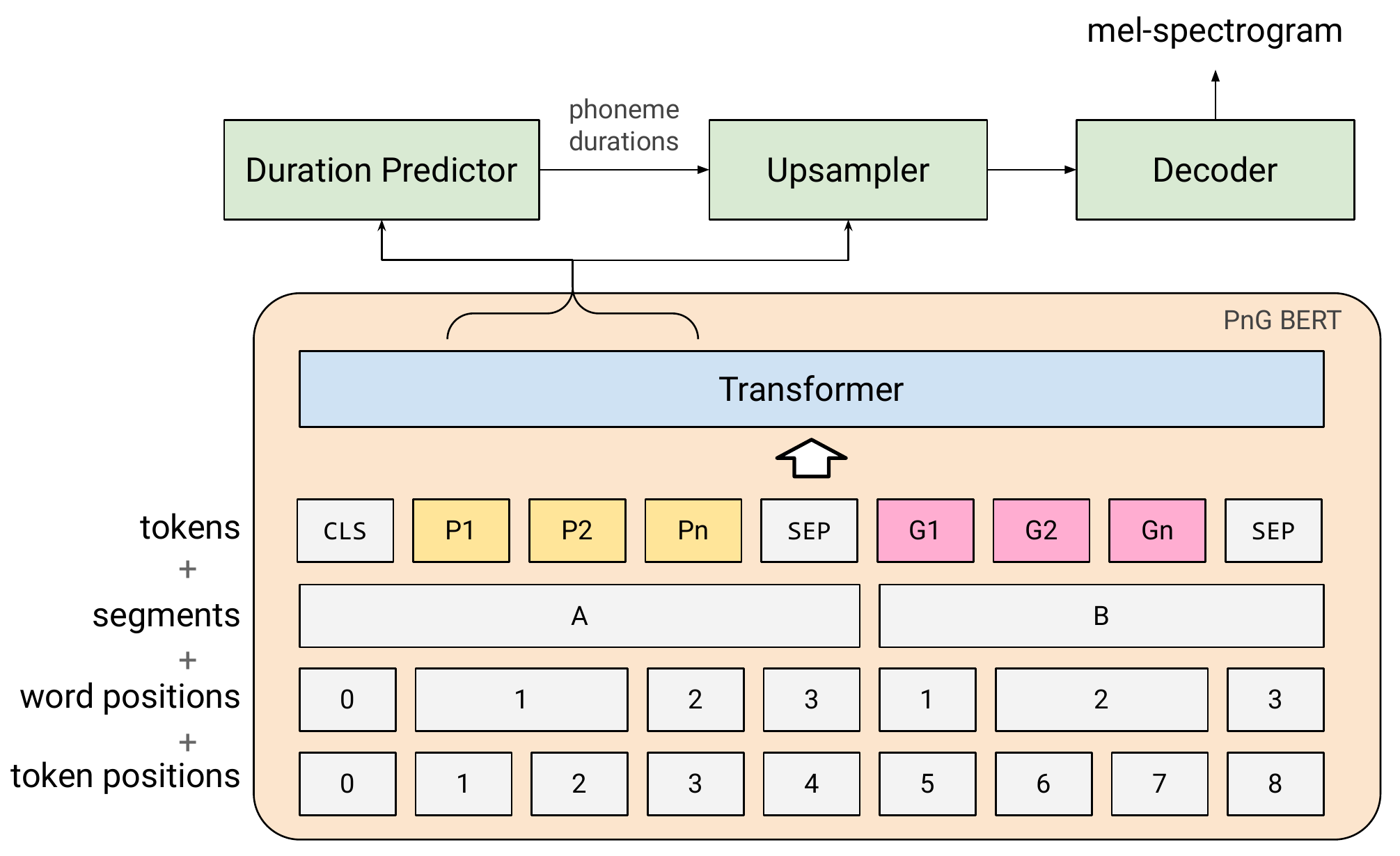}
  \caption{PnG NAT TTS model. The model takes both the phonemes (yellow) and the graphemes (pink) of text as input. PnG BERT can be pre-trained in a self-supervised manner on a large text corpus.}
  \label{fig:png_nat}
  \vspace{-2ex}
\end{figure}

PnG NAT (Figure~\ref{fig:png_nat}) is a combination of PnG BERT \citep{jia2021png} and Non-Attentive Tacotron (NAT) \citep{shen2020non}. It synthesizes speech as natural as professional human speakers \citep{jia2021png}.

PnG BERT is an encoder model specifically designed for neural TTS. It takes both phoneme and grapheme representations of text as input, as well as the word-level alignment between them. %
Similar to BERT \citep{devlin2018bert}, PnG BERT can be pre-trained on a large text corpus in a self-supervised manner.
Experimental results show that PnG NAT
using a pre-trained PnG BERT yields more natural prosody and
more accurate pronunciation than a baseline NAT model using only
phoneme input with no pre-training. Subjective side-by-side (SxS)
preference evaluations show that raters have no statistically significant preference between the speech synthesized using PnG NAT and ground truth studio recordings from professional speakers  \citep{jia2021png}.

We pre-trained PnG BERT on a plain text corpus mined from Wikipedia, containing 131M English sentences, and fine-tuned it in PnG NAT on the entire LibriTTS corpus. We followed the hyperparameters in \citet{jia2021png,shen2020non}.

\paragraph{Performance} 

The performance of the trained PnG NAT model  is evaluated by subjective Mean Opinion Score (MOS, more details in Sec.~\ref{sec:baselines}) on text from LibriTTS test sets in Table~\ref{tbl:tts}.
As can be seen, the synthesized speech obtained about the same naturalness and speaker similarity as the ground truth recordings. 
The self-similarity between different ground truth recordings from this particular speaker is lower than the average on the corpus, reflecting higher expressiveness and more style variation in her recordings. Raters often commented ``lower/higher voice (than the other)'' in the similarity evaluation on the ground truth.

\subsection{PnG NAT with voice cloning}
\label{sec:png-nat-vc}

\begin{figure}[t]
  \centering
  \includegraphics[width=1.0\linewidth]{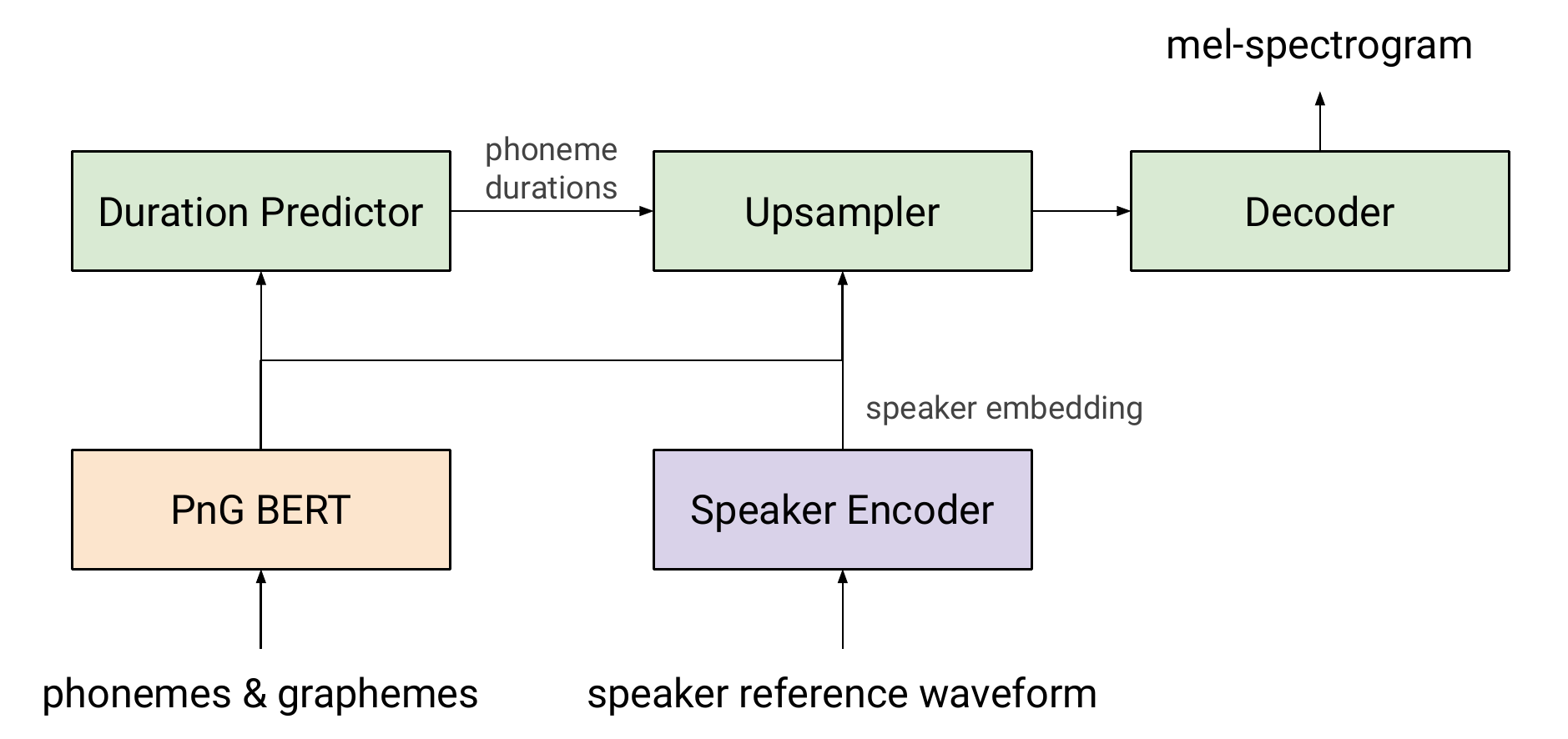}
  \caption{Augmented PnG NAT model for zero-shot voice cloning. The speaker encoder is separately trained in a speaker verification task and is frozen during TTS training.}
  \label{fig:png_nat_d}
  \vspace{-2ex}
\end{figure}

To transfer the voices from the source speech to the translation speech, we modified PnG NAT to support zero-shot cross-lingual voice cloning (VC) by incorporating a speaker encoder in the same way as in \citet{jia2018transfer}. The augmented TTS model is illustrated in Figure~\ref{fig:png_nat_d}. 
The speaker encoder is separately trained in a speaker verification task and is frozen during TTS training. At training time, the paired target speech is used as the reference speech for the speaker encoder. At synthesis time, the phonemes and graphemes in the target language and the reference speech in the source language are fed into the model as inputs, and the model produces speech in the target language with the voice from the  source speech transferred. 

\paragraph{Speaker encoder} 

Compared to the speaker encoder \citep{wan2018generalized} used in \citet{jia2018transfer}, we used an improved model with better performance. The key improvements include:
\begin{inparaenum}[1)]
  \item The model is trained with the generalized end-to-end extended-set softmax loss \citep{pelecanos2021dr};
  \item Instead of LSTM, the model is based on a 256$\times$12 Conformer stack \citep{gulati2020conformer}; 
  \item We introduce an attentive temporal pooling layer \citep{wang2022attentive,pelecanos2022parameter} to aggregate the Conformer output over time, then concatenate the weighted mean and standard deviation, and finally produce the 256-dim speaker embedding with two feed-forward layers.
\end{inparaenum}
This speaker encoder has 21.2M parameters, and is trained on a mixture of a proprietary multilingual speech query dataset covering 37 locales collected by vendors, plus public corpora including LibriVox,
CN-Celeb~\citep{fan2020cn},
TIMIT~\citep{garofolo1993timit},
Fisher~\citep{cieri2004fisher},
and Mixer 4 and 5 \citep{cieri2007resources,brandschain2008speaker}.
The training data contain 122M utterances from 240K speakers in total.
Compared to the speaker encoder used in \citet{jia2018transfer}, the speaker verification Equal Error Rate (EER) on LibriSpeech is reduced from 2.5\% to 0.9\%.

\begin{table}[t]
\centering
\footnotesize
\caption{Subjective MOS of TTS models used for constructing CVSS, evaluated with the CVSS-C target speaker (``3983'') and 500/67 seen/unseen speakers from LibriTTS train/test sets (clean and other), randomly paired with text from LibriTTS test-clean set.}
\vspace{1ex}
\setlength{\tabcolsep}{0.5em}
\begin{tabular}{lccc}
    \toprule
    & Speakers & Naturalness & Similarity \\
    \midrule
    PnG NAT & 3983 & 4.60 $\pm$ 0.06 & 3.77 $\pm$ 0.10 \\
    \midrule
    \multirow{2}{*}{PnG NAT w/ VC} 
    & Seen & 4.01 $\pm$ 0.07 & 3.19 $\pm$ 0.08 \\
    & Unseen & 4.04 $\pm$ 0.07 & 3.00 $\pm$ 0.08 \\
    \midrule
    \multirow{3}{*}{Ground truth} 
    & 3983 & 4.60 $\pm$ 0.06 & 3.87 $\pm$ 0.10 \\
    & Seen & 4.32 $\pm$ 0.06 & 4.30 $\pm$ 0.07 \\
    & Unseen & 4.18 $\pm$ 0.07 & 4.30 $\pm$ 0.06\\
    \bottomrule
\end{tabular}
\vspace{-2ex}
\label{tbl:tts}
\end{table}

\paragraph{Performance} 

The performance of the trained model is evaluated on both seen and unseen speakers from LibriTTS in Table~\ref{tbl:tts}. The synthesized speech obtained high naturalness and speaker similarity, although lower than ground truth recordings, due to the challenge of zero-shot voice transferring, especially when the reference audios are noisy.

\section{Data generation}

\subsection{Data filtering}
\label{s:filtering}

The CoVoST~2 corpus includes a few empty audio files (0 byte) originating from Common Voice version 4. We excluded these audios from CVSS. In addition, we used a proprietary voice activity detector (VAD) to filter out audios without any human voice.
They in total filtered out 133 recordings from CoVoST~2.

\subsection{Text normalization}

We normalize the translation text from CoVoST~2 using a proprietary weighted finite state transducer (WFST)-based text normalizer \citep{ebden2015kestrel}. Non-standard words \citep{sproat2001normalization}, such as numbers, currency expressions, dates, common abbreviations, acronyms, etc., are detected and verbalized. Such normalized text is used as the input 
for TTS synthesis.

For S2ST model training and evaluation, we further converted the normalized text into lowercase, and removed punctuation marks except for apostrophes. This version of the normalized translation text is released in CVSS.
Appendix~\ref{a:textnorm} includes examples of such text normalization.

\begin{table*}[t]
\centering
\footnotesize
\caption{Basic statistics on CVSS-C, CVSS-T and CoVoST 2$^\dagger$ for X$\to$En speech translation. The source languages are sorted by the hours of source recordings in the train sets. ($^\dagger$ Data filtering in Sec~\ref{s:filtering} applied.)}
\vspace{1.5ex}
\begin{tabular}{c|rrr|rrr|rrr|rrr}
    \toprule
    \multirow{4}{*}{X} & \multicolumn{3}{c}{\multirow{2.5}{*}{\#utts}} &  \multicolumn{9}{|c}{Hours} \\
    \cmidrule(l{0.3em}r{0.3em}){5-13}
     & & & & \multicolumn{3}{c|}{CoVoST 2$^\dagger$ (X)} & \multicolumn{3}{c|}{CVSS-C (En)} & \multicolumn{3}{c}{CVSS-T (En)} \\
    \cmidrule(l{0.3em}r{0.3em}){2-13}
      & Train & Dev & Test & Train & Dev & Test & Train & Dev & Test & Train & Dev & Test \\
     \midrule
fr & 207,364 & 14,759 & 14,759 & 264.3 & 21.7 & 23.3 & 174.0 & 13.0 & 13.3 & 192.7 & 14.6 & 15.0 \\
de & 127,822 & 13,511 & 13,504 & 184.3 & 20.7 & 21.5 & 112.4 & 12.5 & 12.1 & 124.2 & 13.6 & 13.4 \\
ca &  95,852 & 12,730 & 12,730 & 135.6 & 19.0 & 20.2 &  88.1 & 12.0 & 12.0 &  95.0 & 12.9 & 13.0 \\
es &  79,012 & 13,212 & 13,216 & 113.1 & 21.8 & 22.7 &  69.5 & 12.4 & 12.4 &  73.7 & 13.2 & 13.3 \\
fa &  53,901 &  3,440 &  3,425 &  49.2 &  4.6 &  5.0 &  25.3 &  2.2 &  2.4 &  29.3 &  2.5 &  2.7 \\
it &  31,698 &  8,938 &  8,951 &  44.2 & 14.3 & 15.4 &  29.4 &  8.5 &  8.6 &  30.5 &  9.2 &  9.5 \\
ru &  12,112 &  6,110 &  6,300 &  18.2 &  9.9 & 10.6 &  13.3 &  6.7 &  6.9 &  13.2 &  6.9 &  7.3 \\
zh &   7,085 &  4,843 &  4,897 &  10.4 &  7.9 &  8.2 &   8.7 &  6.0 &  5.8 &   9.3 &  6.5 &  6.3 \\
pt &   9,158 &  3,318 &  4,023 &  10.3 &  4.4 &  5.3 &   5.7 &  2.1 &  2.6 &   6.5 &  2.4 &  2.9 \\
nl &   7,108 &  1,699 &  1,699 &   7.3 &  1.9 &  2.0 &   4.9 &  1.2 &  1.2 &   5.1 &  1.3 &  1.3 \\
tr &   3,966 &  1,623 &  1,629 &   4.1 &  1.8 &  2.0 &   3.0 &  1.2 &  1.2 &   3.1 &  1.3 &  1.3 \\
et &   1,782 &  1,576 &  1,571 &   3.4 &  2.8 &  2.8 &   2.8 &  2.3 &  2.2 &   2.7 &  2.2 &  2.2 \\
mn &   2,067 &  1,760 &  1,759 &   3.0 &  2.6 &  2.8 &   1.9 &  1.6 &  1.6 &   2.1 &  1.8 &  1.8 \\
lv &   2,337 &  1,125 &  1,629 &   2.1 &  1.1 &  1.7 &   1.2 &  0.6 &  0.8 &   1.4 &  0.7 &  1.0 \\
ar &   2,283 &  1,758 &  1,693 &   2.1 &  1.9 &  1.8 &   1.1 &  0.8 &  0.8 &   1.2 &  1.0 &  0.9 \\
sl &   1,843 &    509 &    360 &   2.0 &  0.5 &  0.4 &   1.1 &  0.3 &  0.2 &   1.3 &  0.4 &  0.2 \\
sv &   2,160 &  1,349 &  1,595 &   1.7 &  1.1 &  1.5 &   1.0 &  0.6 &  0.7 &   1.2 &  0.7 &  0.9 \\
cy &   1,241 &    688 &    690 &   1.7 &  0.9 &  1.0 &   0.9 &  0.5 &  0.5 &   1.0 &  0.5 &  0.5 \\
ta &   1,358 &    384 &    786 &   1.6 &  0.5 &  1.0 &   0.9 &  0.3 &  0.5 &   1.1 &  0.3 &  0.6 \\
ja &   1,119 &    634 &    684 &   1.3 &  0.8 &  0.9 &   0.8 &  0.4 &  0.5 &   0.8 &  0.5 &  0.5 \\
id &   1,243 &    792 &    844 &   1.2 &  0.9 &  0.9 &   0.7 &  0.4 &  0.5 &   0.7 &  0.5 &  0.5 \\
    \midrule
    Total & 652,511	& 94,758 & 96,744 & 861.1 & 141.1 & 151.0 & 546.7 & 85.6 & 86.8 & 596.1 & 93.0 & 95.1 \\
    \bottomrule
\end{tabular}
\label{tbl:stats}
\vspace{-2ex}
\end{table*}

\subsection{TTS synthesis}

\paragraph{CVSS-C} is synthesized using the PnG NAT model described in Sec.~\ref{sec:png-nat}. 
A female speaker ``lavocedorata'' (ID 3983) 
from LibriTTS is used as the canonical speaker. Although this speaker has merely
6.7 minutes recordings in the training set, these recordings are highly fluent, clean and natural. %

\paragraph{CVSS-T} is synthesized using the augmented PnG NAT model described in Sec.~\ref{sec:png-nat-vc} for cross-lingual voice cloning. The speaker embedding computed on the source non-English speech is used for synthesizing the English translation speech.

\paragraph{Vocoder} A neural vocoder based on WaveRNN \citep{kalchbrenner2018efficient} is used for converting the mel-spectrograms synthesized by the TTS models into waveforms. This neural vocoder is trained on a proprietary dataset of 420 hours studio recordings from 98 professional speakers in 6 English accents.

\paragraph{Data format} The synthesized speech is stored as monophonic WAV files at 24~kHz sample rate and in 16-bit linear PCM format.

\subsection{Dataset splitting}

Both CVSS-C and CVSS-T are split into train, dev and test subsets consistently with CoVoST 2.
CoVoST 2 uses an extended CoVoST split in order to increase data utilization from the raw Common Voice dataset, by allowing multiple versions of recordings on the same sentences (likely from different speakers). This extended split is used for the train set of CoVoST~2, while the original Common Voice split is used for the dev and test sets to avoid skew to  duplicate sentences in evaluation. We follow the same data split settings in CVSS.

\section{Statistics}

Basic statistics on both versions of CVSS are shown in Table~\ref{tbl:stats}.
As can be seen, the synthesized translation speech is significantly shorter than the source speech, which is the result of better fluency and the absence of long silences. The duration of CVSS-C is slightly shorter than CVSS-T, indicating faster speaking pace.

\begin{figure}[t]
  \centering
  \includegraphics[width=0.95\linewidth]{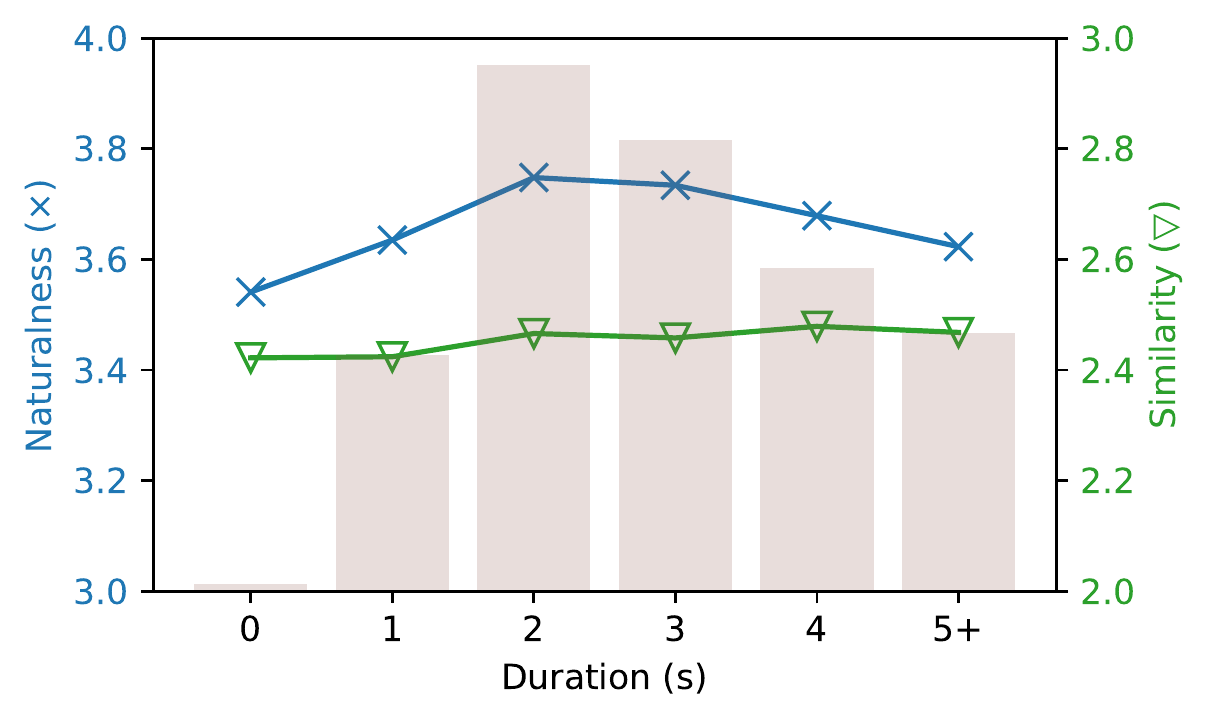}\vspace{-1ex}
  \caption{Naturalness and speaker similarity MOS on CVSS-T breaking down by duration of the translation speech. Bars are the distribution of the durations.}
  \label{fig:breakdown}
  \vspace{-0.6ex}
\end{figure}

The quality of the produced corpus is evaluated as ``Targets'' rows in
Table~\ref{tbl:s2st} and Appendix~\ref{a:s2st-details}.
CVSS-C obtained very high naturalness, while the naturalness and speaker similarity from CVSS-T is lower. Rater comments revealed that the naturalness of CVSS-T is primarily impacted by ``noise'' and ``distortion'', which is likely the result of noisy reference speech from CoVoST~2 used for voice transferring; the speaker similarity is largely impacted by ``different languages'', which does not necessarily reflect voice difference (same as observed in \citet{zhang2019learning,jia2021translatotron}). Objective d-vector similarity on CVSS-T obtained a very high 0.65 despite of the language difference (compared to 0.64 from LibriTTS unseen speakers in a same language in Table~\ref{tbl:tts}), suggesting high speaker similarity estimated for speaker verification. We further break down the evaluation on CVSS-T by speech duration in Figure~\ref{fig:breakdown}, to reflect the impact of the amount of reference audio used for voice cloning. 

Despite the naturalness difference between CVSS-T and CVSS-C, they both obtain high intelligibility, as reflected by ASR BLEU. The ASR BLEU is significantly lower on certain languages (e.g. zh) than others, because those data include a lot of non-English names and proper nouns,  which cannot be recognized correctly by the English ASR model used in evaluation.

\section{Baseline models}
\label{sec:baselines}

On each version of CVSS, we trained two baseline direct S2ST models (Translatotron and Translatotron~2) as well as a baseline cascade S2ST model (ST$\to$TTS).
All models are implemented using the Lingvo framework \citep{shen2019lingvo}.

Following \citet{jia2019direct}, we evaluated the translation quality and speech generation quality of S2ST models. The translation quality is measured by BLEU on ASR transcription from the translation speech (in lowercase, excluding punctuation marks) against the normalized reference translation. Because ASR makes errors, such BLEU can be thought a lower bound of the translation quality. We used an ASR model from \citet{park2020improved} trained on LibriSpeech and LibriLight \citep{kahn2020libri}, and computed BLEU using SacreBLEU \citep{post2018call} with its default configuration.
The speech generation quality is measured subjectively by 5-point mean opinion score (MOS) on naturalness and speaker similarity \citep{jia2018transfer}. Each MOS evaluation was conducted with 1,000 or more ratings by native North American English speakers. Each rater was limited to rate no more than 6 items per evaluation. 

We group the evaluation results on high-resource source languages (French, German, Catalan and Spanish) and low-resource ones (all the rest). MOS evaluation was only conducted on the high-resource language pairs, because otherwise the low translation quality on low-resource languages would negatively impact the subjective assessment of the speech generation quality.

\subsection{Direct S2ST baselines}

\begin{table*}[t]
\centering
\begin{small}
\caption{Multilingual X$\to$En S2ST performance. BLEU is reported as the average on all the 21 (All) and the 4/17 high/low resource (Hi-Res/Lo-Res) language pairs. MOS is evaluated on the 4 high resource language pairs.}
\vspace{1ex}
\begin{tabular}{l|l|cc|rrr}
    \toprule
    \multirow{2.5}{*}{Corpus} & \multirow{2.5}{*}{System} & \multicolumn{2}{c|}{MOS (Hi-Res)} & \multicolumn{3}{c}{BLEU} \\
    \cmidrule{3-7}
     & & Naturalness & Similarity & All & Hi-Res & Lo-Res \\
    \midrule
    \multirow{4.5}{*}{CVSS-C}
    & Translatotron          & 4.29 $\pm$ 0.07 & -- &  3.4 & 11.9 & 1.4 \\
    & Translatotron 2        & 4.61 $\pm$ 0.05 & -- &  8.7 & 25.4 & 4.8 \\
    & Cascade (ST $\to$ PnG NAT) & 4.64 $\pm$ 0.04 & -- & 10.6 & 28.8 & 6.3 \\
    \cmidrule{2-7}
    & Targets                & 4.63 $\pm$ 0.05 & -- & 91.1 & 88.4 & 91.7 \\
    \midrule
    \multirow{4.5}{*}{CVSS-T}
    & Translatotron     & 2.91 $\pm$ 0.09 & 2.34 $\pm$ 0.07 & 4.6 & 16.4 & 1.9 \\
    & Translatotron 2   & 3.80 $\pm$ 0.07 & 2.24 $\pm$ 0.07 & 8.6 & 25.6 & 4.6 \\
    & Cascade (ST $\to$ PnG NAT w/ VC) & 3.66 $\pm$ 0.08 & 2.38 $\pm$ 0.07 & 10.5 & 28.7 & 6.3 \\
    \cmidrule{2-7}
    & Targets           & 3.73 $\pm$ 0.08 & 2.35 $\pm$ 0.07 & 89.9 & 87.1 & 90.5 \\
    \bottomrule
\end{tabular}
\label{tbl:s2st}
\end{small}
\vspace{-2ex}
\end{table*}

On each version of CVSS,
we trained two baseline end-to-end direct S2ST models following Translatotron \citep{jia2019direct} and Translatotron 2 \citep{jia2021translatotron}. For both models, we followed the hyper-parameters from Sec. 5.5 in \citet{jia2021translatotron} except for a few changes. Notably, we used a wider Conformer encoder (256$\times$16) for the larger and more diverse training data. The detailed hyper-parameters are  available in Table~\ref{tbl:hparams}.
All models were trained with a batch size of 768 for 240K steps. We picked checkpoints by the best average BLEU on the dev sets, and report the performance on the test sets in
Table~\ref{tbl:s2st} (detailed in Appendix~\ref{a:s2st-details}).

\subsection{Cascade S2ST baselines}
\label{sec:cascade}

To construct cascade S2ST baselines, we trained an ST model on the original CoVoST~2 corpus, and connected it to the same two TTS models used for constructing CVSS. Note that these cascade models have a data advantage over the direct models at training time (i.e. access to high quality TTS data).

\newrobustcmd{\B}{\bfseries}

\begin{table*}[t]
\centering
\begin{small}
\caption{BLEU of the multilingual X$\to$En ST model, evaluated on the CoVoST~2 test sets. Our model outperforms the previous state-of-the-arts ({\B bold}) trained on CoVoST~2 without using extra data,
and even a few previous works using pre-training with large-scale extra data.
Detailed results are available in Appendix~\ref{a:st-details}.
}
\vspace{1ex}
\setlength{\tabcolsep}{0.6em}
\begin{tabular}{l|cc|rrr|rrrr}
    \toprule
    Model & \#Params & Pre-training & All & Hi-Res & Lo-Res & fr & de & ca & es \\
    \midrule
    \citet{li2020multilingual} (Scratch-BL) & -- & & -- & 14.8 & -- & 24.3 &  8.4 & \B 14.4 & 12.0 \\
    \citet{wang2020covost2} (A2A-L)  & -- & & 5.9 & \B 16.6 & 3.4 & -- & -- & -- & -- \\
    \citet{vyas2021optimally} (base) & 16M & & -- & -- & -- & 22.8 & \B 12.7 & -- & 21.4 \\
    \citet{vyas2021optimally} (deep) & 25M & & -- & -- & -- & \B 25.2 & 8.1 & -- & \B 22.5 \\
    \midrule
    \citet{wang2020covost2} (A2A-L)         & -- & \checkmark & \B 7.5 & 24.0 & \B 3.7 & -- & -- & -- & -- \\
    \citet{wang2020covost2} (A2E-M, arXiv)  & -- & \checkmark &  -- & \B 24.5 &   --  & 27.0 & 18.9 & \B 23.9 & \B 28.0 \\
    \citet{vyas2021optimally} (deep) & 25M & \checkmark & -- & -- & -- & \B 27.3 & \B 20.0 & -- & 25.8 \\
    \midrule
    Ours                     & 43M &            & 11.0 & 29.4 & 6.7 & 31.9 & 23.9 & 27.9 & 33.9 \\
    Ours + ASR pre-training  & 51M & \checkmark & \B 13.3 & \B 31.4 & \B 9.0 & \B 33.8 & \B 26.4 & \B 29.9 & \B 35.6\\
    \midrule
    \multicolumn{9}{l}{\emph{Selected previous works using pre-training on large-scale extra speech, text, and MT data (not including SOTA)}} \\
    XMEF-En \citep{li2020multilingual} & 793M & extra data & 12.4 & 32.4 & 7.7 & \\
    XLS-R (0.3B) \citep{babu2021xls}  & 776M & extra data & 13.2 & 30.6 & 9.2 & \\
    \bottomrule
\end{tabular}
\label{tbl:st}
\end{small}
\vspace{-2ex}
\end{table*}

\paragraph{ST model}

We trained an ST model on the original CoVoST~2 corpus, using the same encoder and decoder architecture and hyper-parameters as in Translatotron~2, except that it predicts 8,192 SentencePiece \citep{kudo2018sentencepiece} tokens
with a beam size of 8, and was trained with a larger batch size and a higher learning rate (Table~\ref{tbl:hparams}). This ST model outperforms the previous state-of-the-art ST models trained on CoVoST~2 without extra data by 5.8 or 6.9 BLEU, as average on all 21 or the 4 high-resource language pairs. It even outperforms a few previous works using models more than 15$\times$ larger, and pre-trained with extra large-scale speech, text, and MT data (although behind even larger ones).  
See Table~\ref{tbl:st} for the performance of this
ST model and Appendix~\ref{a:st-details} for more details.
Such improvements over the previous works partially come from the using of a deeper Conformer encoder which learns better speech representation, and we also noted that the extra regularization on the decoder was crucial for avoiding overfitting.

\subsection{Pre-training}

We explored utilizing pre-training in ASR and ST tasks to improve the performance of both direct and cascade S2ST models. Such pre-training was conducted within the CoVoST~2 corpus without using extra datasets.

Following \citet{wang2020covost2}, we pre-trained a multilingual ASR model on all the 22 languages in CoVoST~2, and used it for initializing the ST models for cascade S2ST. These ASR and ST models used the same model architecture and hyper-parameters as in Sec.~\ref{sec:cascade} except for using a larger 16k multilingual SentencePiece vocabulary.
Similarly, we used the trained ST models to initialize the encoder and decoder of the Translatotron~2 direct S2ST models.

For the simplicity and self-containedness as baselines, we did not explore self-supervised pre-training with extra data in this work. However, such an approach remains promising for improving the performance of S2ST.

\begin{table}[t]
\centering
\footnotesize
\caption{S2ST BLEU on CVSS-C when part of the model is pre-trained in ASR/ST tasks on CoVoST~2. Row 4/5 means initializing the encoder and decoder of Translatotron 2 from the ST models used in row 1/2.}
\vspace{1ex}
\begin{tabular}{lrrr}
    \toprule
    & All & Hi-Res & Lo-Res \\
    \midrule
    Cascade (ST $\to$ TTS) & 10.6 & 28.8 & 6.3 \\
    \quad$+$ ASR pre-training & 12.7 & 30.6 & 8.5 \\
    \midrule
    Translatotron 2 & 8.7 & 25.4 & 4.8 \\
    \quad$+$ ST pre-training & 10.5 & 28.8 & 6.2 \\
    \quad\quad$+$ ASR pre-training & 12.0 & 29.7 & 7.8 \\
    \bottomrule
\end{tabular}
\vspace{-1ex}
\label{tbl:pretraining}
\end{table}

\subsection{Results}

\paragraph{CVSS-C}

As can be seen from Table~\ref{tbl:s2st}, both the cascade model and the Translatotron~2 direct S2ST model produced translation speech as natural as the reference targets, all of which were as natural as human recordings (Table~\ref{tbl:tts}) -- Thanks to the duration-based autoregressive speech generation \citep{shen2020non} used in both the PnG NAT TTS model and the Translatotron~2 S2ST model.
Both of them also obtained translation quality comparable to the ST evaluation (Table~\ref{tbl:st}), with the cascade model performed slightly better, indicating the effectiveness of both cascade and direct S2ST.
The performance of the original Translatotron was behind Translatotron~2 and the cascade S2ST model.

\paragraph{Pre-training (CVSS-C)}

Similarly to observed in ST tasks \citep{weiss2017sequence,bansal2019pre,jia2019leveraging,wang2020covost2}, pre-training with weakly supervised data can benefit the performance of the more difficult task. ASR pre-training further improved the performance of our very strong ST model (Table~\ref{tbl:st}), which in turn led to better performance of the cascade S2ST (Table~\ref{tbl:pretraining}). ST pre-training improved the performance of the  Translatotron~2 direct S2ST models, to be very close to the cascade S2ST models (with 0.1 / 0.7 BLEU differences as average on all language pairs, when initialized from matching ST models without / with ASR pre-training, Table~\ref{tbl:pretraining}).

\paragraph{CVSS-T}

All three models trained on CVSS-T were able to preserve source speakers' voices during speech translation, with about the same speaker similarity to the source speech as the reference targets. 
Similar to the results on CVSS-C, both Translatotron 2 and the cascade model obtained about the same naturalness as the reference targets, with the original Translatotron behind them.
Both Translatotron~2 and the cascade model also obtained ASR BLEU similar to the same on CVSS-C, indicating the effectiveness of Translatotron~2 as a direct S2ST model capable of voice preservation (the performance of the cascade model is expected since it is consistent with the CVSS-T data construction), as well as the high intelligibility of the translation speech in CVSS-T despite of lower naturalness compared to CVSS-C.
Interestingly, the translation quality from the original Translatotron was better on the apparently more difficult CVSS-T dataset than on CVSS-C. This may be explained by the extra task of voice transferring that encouraged its decoder to utilize the attention output. As a matter of fact, inability to pick up attention output is one of the challenges in the original Translatotron tuning \citep{jia2019direct}.

\subsection{Discussion}

Although the translation quality from the direct S2ST models did not surpass the cascade models in our experiments, we observed cases where direct S2ST demonstrated advantages over the latter, in terms of avoiding error propagation on rare words, which is a known challenge for ST \citep{gaido2021moby}. For example, for a German source speech with content ``Mogadischu ist die Hauptstadt von Somalia'', the ST model in the cascade S2ST mistakenly translated the speech corresponding to ``Mogadischu'' into English text as ``UgoDIShu'', which turned into being considered as four words by the downstream TTS model because of text normalization, and finally produced translation speech unable to be understood (ASR transcribed it into ``hugo d i shoo'', with ``d'' and ``i'' pronounced as individual letters). As a comparison, the Translatotron~2 direct S2ST model mostly copied the pronunciation from the source speech into the translation speech. Although it was not able to be recognized correctly by the ASR model for evaluation (transcribed as ``bogodisu''), it was able to be understood by humans.
Similar examples were reported in \citet{jia2019direct}. This can be a potential advantage of direct S2ST worth further exploration.

\section{Conclusion}

We described two massively multilingual-to-English S2ST datasets, \emph{CVSS-C} and \emph{CVSS-T}, each with about 1.9K hours of sentence-level parallel S2ST pairs, covering 21 source languages. The translation speech in CVSS-C is in a single canonical speaker's voice, while the same in CVSS-T is in voices transferred from the source speech. Each dataset provides unique values not existing in other public S2ST corpora.

We built baseline multilingual direct S2ST models and cascade S2ST models on both datasets, verifying the effectiveness of the corpus. To build strong cascade S2ST baselines, we trained an ST model on CoVoST~2, which outperforms the previous state-of-the-art by 5.8 BLEU. Nevertheless, the performance of the direct S2ST models approaches the strong cascade baselines when trained from scratch, and
with only 0.1 or 0.7 BLEU difference on ASR transcribed translation when %
initialized from matching ST models.

Future work includes expanding the corpus coverage to En$\to$X directions.

\section{Acknowledgements}

We acknowledge the volunteer contributors and the organizers of the Common Voice\footnote{\url{https://commonvoice.mozilla.org/}} and LibriVox \footnote{\url{https://librivox.org/}} projects for their contribution and collection of recordings, the creators of Common Voice \citep{ardila2020common}, CoVoST \citep{wang2020covost}, CoVoST 2 \citep{wang2020covost2}, Librispeech \citep{panayotov2015librispeech} and LibriTTS \citep{zen2019libritts} corpora for their previous works. We would like to thank Ankur Bapna for helps on data processing, Yiling Huang and Jason Pelecanos for improving the speaker encoder model, and Colin Cherry and Alexis Conneau for helpful feedback.

\begin{table}[t!]
\centering
\begin{small}
\caption{Hyper-parameters of the baseline models. (``T1'': Translatotron; ``T2'': Translatotron 2; ``$\times n$'': $n$ layers. $^\dagger$ T2 uses a 128-dim pre-net on CVSS-C and a 16-dim one on CVSS-T, which is crucial for voice preservation.)}
\vspace{1ex}
\setlength{\tabcolsep}{0em}
\begin{tabular}{l@{\hskip -0.6em}ccc}
    \toprule
     & T1 & T2 & ST \\
    \midrule
    \multicolumn{4}{l}{\emph{Input}} \\
    Sample rate (Hz)    & \multicolumn{3}{c}{48,000} \\
    Mel channels        & \multicolumn{3}{c}{80} \\
    Lower / Upper band (Hz) & \multicolumn{3}{c}{125 / 7,600} \\
    Frame size / step (ms)     & \multicolumn{3}{c}{25 / 10} \\
    \midrule
    \multicolumn{4}{l}{\emph{Output}} \\
    Sample rate (Hz)    & \multicolumn{2}{c}{24,000} & \multirow{4}{*}{--} \\
    Mel channels        & \multicolumn{2}{c}{128} \\
    Lower / Upper band (Hz) & \multicolumn{2}{c}{20 / 12,000} \\
    Frame size / step (ms)     & \multicolumn{2}{c}{50.0 / 12.5} \\
    \midrule
    \multicolumn{4}{l}{\emph{SpecAugment}} \\
    Freq / Time blocks & \multicolumn{3}{c}{2 / 10} \\
    Freq / Time length ratio & \multicolumn{3}{c}{0.33 / 0.05} \\
    \midrule
    \multicolumn{4}{l}{\emph{Encoder}} \\
    Conformer dims   & \multicolumn{3}{c}{256 $\times$ 16} \\
    Attention heads  & \multicolumn{3}{c}{4} \\
    Conv kernal size & \multicolumn{3}{c}{32} \\
    Subsample factor & \multicolumn{3}{c}{4}\\
    \midrule
    \multicolumn{4}{l}{\emph{Attention}} \\
    Output dim      & 256 & \multicolumn{2}{c}{256} \\
    Hidden dim      & 1024 & \multicolumn{2}{c}{512} \\
    Attention heads & 4 & \multicolumn{2}{c}{8} \\
    Dropout prob    & 0.1 & \multicolumn{2}{c}{0.1} \\
    \midrule
    \multicolumn{4}{l}{
    \emph{Decoder (Auxiliary decoder in T1)}} \\
    LSTM dims    & 256 $\times$ 2 & \multicolumn{2}{c}{512 $\times$ 4} \\
    Zoneout prob & 0.1 & \multicolumn{2}{c}{0.1}\\
    Per-layer dropout  & 0.2 & \multicolumn{2}{c}{0.2}\\
    Embedding dim & 96 & \multicolumn{2}{c}{128} \\
    Label smoothing  & 0.0 & \multicolumn{2}{c}{0.1} \\
    Loss weight & 1.0 & 10.0 & 1.0 \\
    \midrule
    \multicolumn{4}{l}{\emph{Duration predictor}} \\
    Bi-LSTM dims & \multirow{2}{*}{--} & 128 $\times$ 2 & \multirow{2}{*}{--}\\
    Loss weight & & 1.0 & \\
    \midrule
    \multicolumn{4}{l}{
    \emph{Synthesizer (Spectrogram decoder in T1)}} \\
    LSTM dims  & 1024 $\times$ 6 & 1024 $\times$ 2 & \multirow{6}{*}{--}\\
    Zoneout prob & 0.1 & 0.1 & \\
    Pre-net dim & 32  & 128 / 16 $^\dagger$  & \\
    Pre-net dropout & \multicolumn{2}{c}{0.5} & \\
    Post-net  &  \multicolumn{2}{c}{(5, 512) $\times$ 4  $+$ (5, 128)} \\
    Loss weight & \multicolumn{2}{c}{1.0} \\
    \midrule
    \multicolumn{4}{l}{\emph{Training (Learning rate schedule from \citet{vaswani2017attention})}} \\
    Learning rate (peak) & \multicolumn{2}{c}{1.33$\times10^{-3}$} & 3.13$\times10^{-3}$  \\
    Warm-up steps & \multicolumn{2}{c}{20K} & 10K \\
    Batch size & \multicolumn{2}{c}{768} & 2,048 \\
    Training steps & \multicolumn{2}{c}{240K} & 100K \\
    \bottomrule
\end{tabular}
\label{tbl:hparams}
\end{small}
\vspace{-2ex}
\end{table}

\clearpage

\if\arxiv1
\renewcommand{\bibsection}{\section{Bibliographical References}}
\bibliographystyle{plainnat}
\else
\section{Bibliographical References}
\bibliographystyle{lrec2022-bib}
\fi
\bibliography{references}

\newpage
\begin{appendices}
\appendix
\onecolumn

\section{Examples of normalized text}
\label{a:textnorm}

\begin{table*}[h]
\centering
\footnotesize
\caption{Examples of normalized translation text.}
\vspace{1.5ex}
\begin{tabular}{ll}
    \toprule
    Key & \texttt{common\_voice\_fr\_17557881.mp3} \\
    Original & \texttt{I was given three identical amendments, numbers 20, 59 and 132.} \\
    Normalized & \makecell[tl]{\texttt{i was given three identical amendments numbers twenty fifty nine and one}\\ \texttt{hundred thirty two}} \\
    \midrule
    Key & \texttt{common\_voice\_fr\_19176154.mp3} \\
    Original & \texttt{The musical genre of the song is 100\% Disco.} \\
    Normalized & \texttt{the musical genre of the song is one hundred percent disco} \\
    \midrule
    Key & \texttt{common\_voice\_fr\_17939186.mp3} \\
    Original & \texttt{Believe me, Tyroleans, God is with us! Mulhdorf, 27 April 1809.} \\
    Normalized & \makecell[tl]{\texttt{believe me tyroleans god is with us mulhdorf the twenty seventh of april}\\ \texttt{eighteen o nine}} \\
    \midrule
    
    Key & \texttt{common\_voice\_fr\_17861547.mp3} \\
    Original & \texttt{28 boulevard Henri Sizaire, 81100 Castres.}  \\
    Normalized & \texttt{twenty eight boulevard henri sizaire eight one one o o castres} \\
    \midrule
    Key & \texttt{common\_voice\_fr\_17558962.mp3} \\
    Original & \texttt{That is why the RRDP group supports this proposition of law.} \\
    Normalized & \texttt{that is why the r r d p group supports this proposition of law} \\
    \midrule
    Key & \texttt{common\_voice\_de\_18737961.mp3} \\
    Original & \texttt{You can't go through a 30s zone with 70!}  \\
    Normalized & \texttt{you can't go through a thirties zone with seventy} \\
    \midrule
    Key & \texttt{common\_voice\_zh-CN\_18885718.mp3} \\
    Original & \makecell[tl]{\texttt{Prince Frederick, member of British Royal Family, Grandson of King George}\\ \texttt{II, brother of King George III.}} \\
    Normalized & \makecell[tl]{\texttt{prince frederick member of british royal family grandson of king george}\\ \texttt{the second brother of king george the third}} \\
    \midrule
    Key & \texttt{common\_voice\_zh-CN\_19026623.mp3} \\
    Original & \texttt{Youqichuangong(\begin{CJK*}{UTF8}{gbsn}有栖川宫\end{CJK*}), the sixth emperor} \\
    Normalized & \texttt{youqichuangong you qi chuan gong the sixth emperor} \\
    \bottomrule
\end{tabular}
\label{tbl:textnorm}
\end{table*}

\newpage

\section{Detailed performance of the S2ST models}
\label{a:s2st-details}

\begin{table*}[h]
\centering
\begin{small}
\caption{Multilingual X$\to$En S2ST BLEU on the 21 language pairs. Source languages are sorted by the hours of source speech in the train sets (the second header row).}
\vspace{1.5ex}
\setlength{\tabcolsep}{0.6em}

\begin{tabular}{l|l|r|rrrr|rrrrrr}
    \toprule
    \multirow{2.5}{*}{Corpus} & \multirow{2.5}{*}{System} & \multirow{2}{*}{Avg} & fr & de & ca & es & fa & it & ru & zh & pt & nl \\
    \cmidrule(l{0.3em}r{0.3em}){4-13}
    & & & 264 & 184 & 136 & 113 & 49 & 44 & 18 & 10 & 10 & 7\\
    \midrule
    \multirow{7.5}{*}{CVSS-C}
    & Translatotron       &  3.4 & 15.5 & 6.9 & 11.0 & 14.1 & 1.4 & 9.3 & 4.3 & 1.5 & 2.2 & 2.1 \\
    & Translatotron 2     &  8.7 & 28.3 & 19.7 & 23.5 & 30.1 & 2.4 & 24.1 & 19.6 & 4.5 & 12.5 & 6.5 \\
    & \quad  
      $+$ ST pre-training & 10.5 & 31.4 & 23.8 & 26.9 & 32.9 & 3.8 & 27.8 & 21.1 & 6.2 & 12.6 & 12.7 \\
    & \quad\quad  
     $+$ ASR pre-training & 12.0 & 32.4 & 24.8 & 28.2 & 33.4 & 6.3 & 28.6 & 23.2 & 6.3 & 18.3 & 15.8 \\
    & Cascade (ST $\to$ TTS)  & 10.6 & 31.2 & 23.9 & 26.8 & 33.3 & 3.4 & 28.1 & 24.4 & 6.8 & 14.8 & 9.8 \\  
    & \quad  
     $+$ ASR pre-training & 12.7 & 32.9 & 26.2 & 28.6 & 34.9 & 5.6 & 30.2 & 27.1 & 8.7 & 19.8 & 14.4 \\
    \cmidrule{2-13}
    & Targets             & 91.1 & 84.6 & 88.4 & 92.0 & 88.6 & 91.7 & 89.5 & 94.0 & 77.8 & 93.1 & 90.6 \\
    \midrule
    \multirow{4.5}{*}{CVSS-T}
    & Translatotron       & 4.6 & 20.0 & 10.4 & 15.2 & 19.8 & 1.6 & 14.0 &  6.0 & 1.6 & 3.3 & 3.2 \\
    & Translatotron 2     & 8.6 & 28.5 & 19.7 & 23.7 & 30.5 & 2.4 & 24.4 & 18.3 & 5.1 & 9.0 & 7.8 \\
    & Cascade (ST $\to$ TTS w/ VC) & 10.5 & 31.1 & 23.8 & 26.7 & 33.3 & 3.4 & 28.1 & 24.4 & 6.7 & 14.7 & 9.8 \\
    \cmidrule{2-13}
    & Targets             & 89.9 & 83.5 & 86.9 & 91.1 & 87.0 & 90.5 & 88.0 & 92.9 & 76.4 & 92.0 & 89.5 \\
    \bottomrule
\end{tabular}

\begin{tabular}{l|l|rrrrrrrrrrr}
    \toprule
    \multirow{2.5}{*}{Corpus} & \multirow{2.5}{*}{System} & tr & et & mn & ar & lv & sl & sv & cy & ta & ja & id \\
    \cmidrule(l{0.3em}r{0.3em}){3-13}
    & & 4.1 & 3.4 &  3.0 &  2.1 &  2.1 &  2.0 &  1.7 &  1.7 &  1.6 & 1.3 & 1.2 \\
    \midrule
    \multirow{7.5}{*}{CVSS-C}
    & Translatotron       & 1.2 & 0.1 & 0.1 & 0.1 & 0.2 & 0.3 & 0.4 & 0.3 & 0.1 & 0.2 & 0.1 \\
    & Translatotron 2     & 3.8 & 0.6 & 0.2 & 1.7 & 1.5 & 0.4 & 1.3 & 0.9 & 0.1 & 0.5 & 0.4 \\
    & \quad  
      $+$ ST pre-training & 7.5 & 1.5 & 0.3 & 4.0 & 2.4 & 0.9 & 2.0 & 1.4 & 0.1 & 0.5 & 0.5 \\
    & \quad\quad  
     $+$ ASR pre-training & 10.6 & 2.5 & 0.4 & 5.4 & 2.3 & 3.1 & 3.2 & 4.5 & 0.1 & 1.0 & 1.0 \\
    & Cascade (ST $\to$ TTS)  & 5.1 & 1.7 & 0.3 & 4.1 & 2.3 & 0.6 & 1.4 & 2.1 & 0.2 & 0.7 & 0.9 \\
    & \quad  
     $+$ ASR pre-training & 10.7 & 3.2 & 0.6 & 7.8 & 2.8 & 2.0 & 3.4 & 5.0 & 0.2 & 0.9 & 1.6 \\
    \cmidrule{2-13}
    & Targets             & 92.7 & 89.3 & 92.4 & 94.2 & 94.8 & 94.9 & 94.1 & 92.0 & 90.6 & 95.3 & 92.6  \\
    \midrule
    \multirow{4.5}{*}{CVSS-T}
    & Translatotron       & 0.2 & 0.2 & 0.1 & 0.3 & 0.2 & 0.2 & 0.2 & 0.7 & 0.1 & 0.1 & 0.1 \\
    & Translatotron 2     & 4.8 & 0.9 & 0.1 & 1.0 & 0.6 & 0.3 & 0.7 & 1.6 & 0.1 & 0.3 & 0.2  \\
    & Cascade (ST $\to$ TTS w/ VC) & 5.2 & 1.7 & 0.3 & 4.1 & 2.3 & 0.6 & 1.5 & 2.0 & 0.2 & 0.6 & 0.9 \\
    \cmidrule{2-13}
    & Targets             & 91.3 & 87.6 & 90.9 & 93.1 & 93.5 & 93.0 & 93.1 & 91.1 & 89.8 & 94.4 & 91.9 \\
    \bottomrule
\end{tabular}
\label{tbl:s2st-details}
\end{small}
\end{table*}

\section{Detailed performance of the ST models}
\label{a:st-details}

\begin{table*}[h]
\centering
\begin{small}
\caption{Performance of the multilingual X$\to$En ST model used for the cascade S2ST baselines. Evaluated by BLEU on CoVoST 2 test sets.}
\vspace{1.5ex}
\begin{tabular}{l|rrrrrrrrrrr}
    \toprule
     & \multicolumn{1}{c|}{Avg} & fr & de & ca & \multicolumn{1}{r|}{es} & fa & it & ru & zh & pt & nl \\

    \midrule
    Ours                    & \multicolumn{1}{r|}{11.0} & 31.9 & 23.9 & 27.9 & \multicolumn{1}{r|}{33.9} & 3.4 & 27.9 & 25.4 & 8.7 & 15.3 & 10.8 \\
    Ours + ASR Pre-training & \multicolumn{1}{r|}{13.3} & 33.8 & 26.4 & 29.9 & \multicolumn{1}{r|}{35.6} & 5.5 & 29.9 & 28.1 & 10.7 & 20.0 & 15.7 \\
    \bottomrule
    \toprule
    & tr & et & mn & ar & lv & sl & sv & cy & ta & ja & id \\
    \midrule
    Ours                    &  5.2 & 2.1 & 0.4 & 4.5 & 2.6 & 0.7 & 1.8 & 2.6 & 0.2 & 0.7 & 1.2 \\
    Ours + ASR Pre-training & 10.2 & 3.5 & 0.7 & 8.6 & 3.5 & 2.7 & 4.3 & 5.9 & 0.2 & 1.0 & 2.2 \\
    \bottomrule
\end{tabular}
\label{tbl:st-details}
\end{small}
\end{table*}

\end{appendices}

\end{document}